\documentclass[conference, final]{IEEEtran}

\IEEEoverridecommandlockouts
\usepackage{cite}
\usepackage{amsmath,amssymb,amsfonts}
\usepackage{algorithmic}
\usepackage{graphicx}
\usepackage{textcomp}

\usepackage[inline]{enumitem}
\usepackage{xcolor}

\usepackage{witharrows}

\usepackage{siunitx}

\usepackage{siunitx}
\usepackage[flushleft]{threeparttable}
\usepackage{tablefootnote}
\usepackage{array}

\newcolumntype{P}[1]{>{\centering\arraybackslash}m{#1}}
\usepackage{multicol}
\usepackage{multirow}
\usepackage{tabulary}
\usepackage{colortbl}
\definecolor{mygray}{gray}{0.90}
\usepackage{makecell}
\usepackage{booktabs}
\usepackage{subcaption}
\usepackage{stmaryrd}
\usepackage{float}
\usepackage{pifont}

\usepackage{arydshln}
\colorlet{mygray}{gray!15!white}

\usepackage[switch,columnwise]{lineno}

\usepackage{url,hyperref,microtype}
\hypersetup{
    colorlinks=true,
    citecolor=blue,
    linkcolor=blue,
    filecolor=magenta,      
    urlcolor=black,
}

\def\BibTeX{{\rm B\kern-.05em{\sc i\kern-.025em b}\kern-.08em
    T\kern-.1667em\lower.7ex\hbox{E}\kern-.125emX}}
\begin{document}

\title{FOCUS: A Lightweight Mamba-2 Model for EEG-Based Cognitive Workload Classification
\\
}

\author{

\IEEEauthorblockN{Ioannis Kyprakis}
\IEEEauthorblockA{
\textit{Hellenic Mediterranean University} \\
Heraklion, Greece \\
ikyprakis@ics.forth.gr
}

\and

\IEEEauthorblockN{Stefanos Gkikas}
\IEEEauthorblockA{
\textit{Honda Research Institute Japan} \\
Wako City, Japan \\
stefanos.gkikas@jp.honda-ri.com
}

\and

\IEEEauthorblockN{Chariklia Chatzaki}
\IEEEauthorblockA{\textit{Hellenic Mediterranean University} \\
Heraklion, Greece  \\
rchatzaki@hmu.gr}

\and

\IEEEauthorblockN{Zongyi Jiang}
\IEEEauthorblockA{
\textit{University of Canberra} \\
Canberra, Australia \\
zongyi.jiang@canberra.edu.au
}

\and

\IEEEauthorblockN{Randy Gomez}
\IEEEauthorblockA{\textit{Honda Research Institute Japan} \\
Wako City, Japan \\
r.gomez@jp.honda-ri.com}

 \and

\IEEEauthorblockN{Raul Fernandez Rojas}
\IEEEauthorblockA{
\textit{BioSIS (Biosensing \& Intelligent Systems) Lab} \\
\textit{Centre for Intelligent Computing and Systems} \\
\textit{University of Canberra} \\
Canberra, Australia \\
raul.fernandezrojas@canberra.edu.au
}

\and
\IEEEauthorblockN{Manolis Tsiknakis}
\IEEEauthorblockA{
\textit{Hellenic Mediterranean University} \\
Heraklion, Greece \\
tsiknaki@hmu.gr
}

}

\maketitle

\begin{abstract}

Cognitive workload estimation from physiological signals has become more relevant in high-risk environments, where continuous and unobtrusive monitoring can inform adaptive systems and safety interventions. This paper presents
FOCUS (Frequency-band Optimized Cognitive-load Ultralight SSM), a compact Mamba-2-based model for EEG-based binary
workload classification under constrained data conditions.
FOCUS does not learn frequency sensitivity from data. Instead, it encodes preprocessed multi-channel EEG with a fixed FFT decomposition that covers the five canonical EEG bands.
The proposed model combines depthwise-separable convolution and a weight-shared encoder-bottleneck Mamba-2 pair, with 0.04\, M parameters and 0.11\, GFLOPs.
The model was evaluated in a pilot study with 11 participants under leave-one-subject-out cross-validation on three cognitively different tasks, and on the full dataset, FOCUS attains 74.75\% accuracy at a GPU model-forward latency of 0.08\, ms.
Within this controlled pilot setting, the selected FOCUS configuration offers a good trade-off between classification performance and computational cost. In the weight-sharing ablation, the shared configuration achieves 4.55 percentage points higher accuracy on the full dataset than the non-shared variant, while it also reduces the SSM parameter budget; under a two-sided Wilcoxon signed-rank test, this difference is not significant ($p=0.083$).

\end{abstract}


\section{Introduction}

Cognitive workload refers to the mental effort that working memory requires in order to perform a task. Especially in high-risk domains such as aviation, automotive systems, robotics, and automated industries, real-time cognitive workload estimation has important implications for safety and performance. Beyond professional environments, detecting when a person may need support due to increased cognitive workload is also relevant in areas such as health care and education. Early identification of elevated cognitive workload, for instance, may allow timely assistance \cite{hassard_teoh_2018} \cite{anders_moontaha_2024}.
Workload assessment methods are generally divided into subjective and objective approaches \cite{rojas_debie_2020}. Subjective tools, such as the NASA Task Load Index (NASA-TLX), are well validated; however, they can only be applied after task completion, so real-time monitoring is not possible \cite{galy_paxion_2018}. Objective approaches exploit the relationship between mental effort and autonomic or cortical nervous system activity; in this way, workload can be estimated continuously through physiological sensors \cite{charles_nixon_2019}. Common physiological modalities for cognitive workload estimation include electroencephalography (EEG), electrocardiography (ECG), electrodermal activity (EDA), and respiratory signals.
Cognitive demand is multidimensional. Different physiological and neuroimaging signals capture different aspects of it \cite{hirachan_niraj_2022}. The cognitive workload literature has explored a plethora of modality configurations, from focused single-signal approaches to predefined multimodal combinations, and each of them has shown value for different tasks and deployment contexts \cite{hogervorst_brouwer_2014, debie_rojas_2021, charles_nixon_2019, tao_tan_2019}. A significant portion of the multimodal architectures in this area employs modality-specific feature extraction and explicit fusion mechanisms, such as intra-modality encoders, dual-branch networks, or decision-level modules, which are typically designed around a fixed sensor configuration and optimized for specific input settings \cite{hu_sukthankar_2024, rabbani_islam_2024, li_zhu_2025}.

Among these physiological modalities, EEG has received particular attention, since it directly measures cortical activity and captures rapid changes associated with cognitive engagement \cite{gevins_smith_2003}. EEG-based workload studies often examine oscillatory activity, because alpha and theta rhythms have been associated with cognitive and memory processes \cite{klimesch_1999}. For these reasons, EEG is relevant for monitoring workload in applied domains where excessive cognitive demand may affect safety and performance, including aviation, driving, and human-machine interaction \cite{borghini_astolfi_2014}. However, EEG-based workload assessment is difficult. Workload-related neural patterns can vary across individuals, tasks, and experimental contexts, and for this reason, generalization across task conditions is an important issue \cite{ke_qi_2014}.

In this pilot study, $11$ participants followed a controlled experimental protocol and completed three cognitive tasks; a fluid intelligence reasoning task, a numerical processing task, and an interactive video game. Each task was administered at two difficulty levels, easy and hard, in order to induce different levels of cognitive workload, while continuous EEG recordings were acquired during task performance in order to capture the workload-related neural activity of both conditions.
In addition, this study proposes Frequency-band Optimized Cognitive-load Ultralight SSM (FOCUS), a compact Mamba-2-based sequence model designed for high efficiency and low model-side computational cost in adaptive systems and user-support applications.

\section{Related Work}
\label{related_work}

The EEG modality has been studied extensively for cognitive workload estimation. CNN and LSTM architectures are among the most common choices for extracting spatial and temporal representations from raw or preprocessed signals \cite{qin_bulbul_2023}. However, more complex network designs have shown improved performance \cite{siddhad_roy_2024}, while systematic reviews report that, in general, deep learning approaches outperform feature engineering pipelines \cite{khingphai_moshfeghi_2025}. Transformer and Conformer architectures have also been applied to EEG-based workload estimation and attained competitive results at low computational cost \cite{song_zheng_2023, gkikas_cruz_eeite_cwl_2026}.

Beyond engineering and human-machine interaction, EEG-based monitoring of cognitive workload has gained particular relevance in health-related settings. There, cognitive overload and fatigue may affect clinical performance and the safety of patients. As the studies in this area indicate, the use of EEG-based workload estimation extends from performance optimization to adaptive health informatics, medical training, and cognitive monitoring systems, where it may constitute a useful measurement modality. Regarding the surgical domain, the authors in \cite{morales_ruiz-rabelo_2019} showed that prefrontal EEG beta activity differentiates the workload levels imposed by surgical procedures of varying complexity. According to their findings, EEG-based indices can provide objective, nonintrusive measures of mental overload in surgeons and may support the monitoring of surgeons and residents during training. More recently, the study in \cite{atheef_powar_2026} extended this line of research to robot-assisted surgery with machine learning frameworks that estimate cognitive workload from EEG epochs recorded across multiple cortical regions, including the frontal, temporal, parietal, and occipital areas. In simulation-based medical education, the relationship between cognitive fatigue and medical error was examined in \cite{krigolson_howse_2025}, where mobile EEG was proposed as an objective tool for assessing the cognitive fatigue of medical trainees during simulated clinical work. Outside the surgical context, EEG-based passive brain--computer interfaces have been explored for adaptive learning systems \cite{beauchemin_charland_2024}, in which real-time cognitive load estimation can adjust the pace of information presentation dynamically. In cognitive training contexts, wearable EEG setups were shown to be able to track workload during N-back tasks \cite{kutafina_heiligers_2021}. The specific setups could be relevant for evaluating cognitive training approaches, including future clinical or rehabilitation-oriented applications.

\section{Methodology}

This section describes the data acquisition and pre-processing
pipeline, as well as the proposed \textit{FOCUS} architecture
for EEG-based cognitive workload classification.

\subsection{Data Collection}

We employ the EEG dataset of~\cite{gkikas_cruz_eeite_cwl_2026}, in which eleven participants took part (6 male and 5 female; ages 20--40 years; mean age $25\!\pm\!5.5$). On the day of the experiment, no participant reported a neurological disorder or the use of substances that affect the nervous system. The Institutional Review Board approved the study protocol, and every participant gave written informed consent.
Three tasks of different cognitive nature were performed, namely abstract reasoning (IQ, Raven's Progressive Matrices), numerical problem solving (MATH), and an interactive video game (GAME, Flappy Bird).
Each task was presented at two predefined difficulty levels (Easy and Hard). These controlled experimental conditions were designed to induce differences in cognitive demand, and their validity was examined previously with the same task protocol. Analysis of NASA-TLX responses showed significant differences between the Easy and Hard conditions for the Raven and Numerical tasks~\cite{hirachan_niraj_2022}, while an EEG-based mixed-model ANOVA showed significant differences between the Easy and Hard conditions for the Numerical and Game tasks~\cite{mathews_hirachan_2024}. Both the subjective and the EEG-based analyses thus provide prior evidence in favor of the experimental difficulty manipulation.
Continuous EEG was recorded with the \textit{EMOTIV EPOC} wireless headset, which has $C=14$ measurement channels (AF3, F7, F3, FC5, T7, P7, O1, O2, P8, T8, FC6, F4, F8, AF4) placed according to the international 10--20 system, at a sampling frequency of $f_s = 128$\,Hz. Each participant contributed 12 samples per difficulty level and task, i.e., $N=264$ samples per task (EASY\,=\,132, HARD\,=\,132).
Consequently, each held-out LOSO fold is balanced between the EASY and HARD classes, since it contains 12 samples per class for each individual task and 36 samples per class in the pooled ALL setting.
{Each task-difficulty condition lasted 2\,min and was divided into twelve non-overlapping 10\,s samples. One sample therefore corresponds to a complete 10\,s, 14-channel EEG input with $T=1280$ time points.

\subsection{Pre-processing}

The EEG signals were pre-processed offline using EEGLAB in MATLAB. Independent Component Analysis (ICA) was applied in order to remove eye-blink and muscular artifacts. Components associated with non-neural activity were visually inspected and excluded. The outcome consists of clean 10\,s samples with $T=1280$ time points. The offline ICA artifact-removal stage is not included in the reported model-forward latency.

\subsection{Proposed Architecture: FOCUS}

We propose \textit{FOCUS}, a compact sequence model founded on Mamba-2 for preprocessed multi-channel EEG samples. Four explicit decisions shape its design; their aim is a minimal number of parameters, while the discriminative capacity is preserved:

\begin{itemize}
  \item[\textbf{D1:}] Fixed FFT band decomposition (zero
    learnable parameters in the filter stage).
  \item[\textbf{D2:}] Depthwise-separable temporal convolution
    (considerably fewer parameters than a standard
    convolution).
  \item[\textbf{D3:}] Weight sharing between the encoder and
    bottleneck Mamba-2 blocks (halves the SSM parameters).
  \item[\textbf{D4:}] Minimal model dimensionality
    ($d=64$, $d_\mathrm{state}=16$).
\end{itemize}
The model receives a preprocessed multi-channel EEG sample $\mathbf{X} \in \mathbb{R}^{C \times T}$ and outputs a binary workload label.
Figure~\ref{fig:FOCUS} illustrates the full pipeline.

\begin{figure}[t]
    \centering
    \includegraphics[width=\columnwidth]{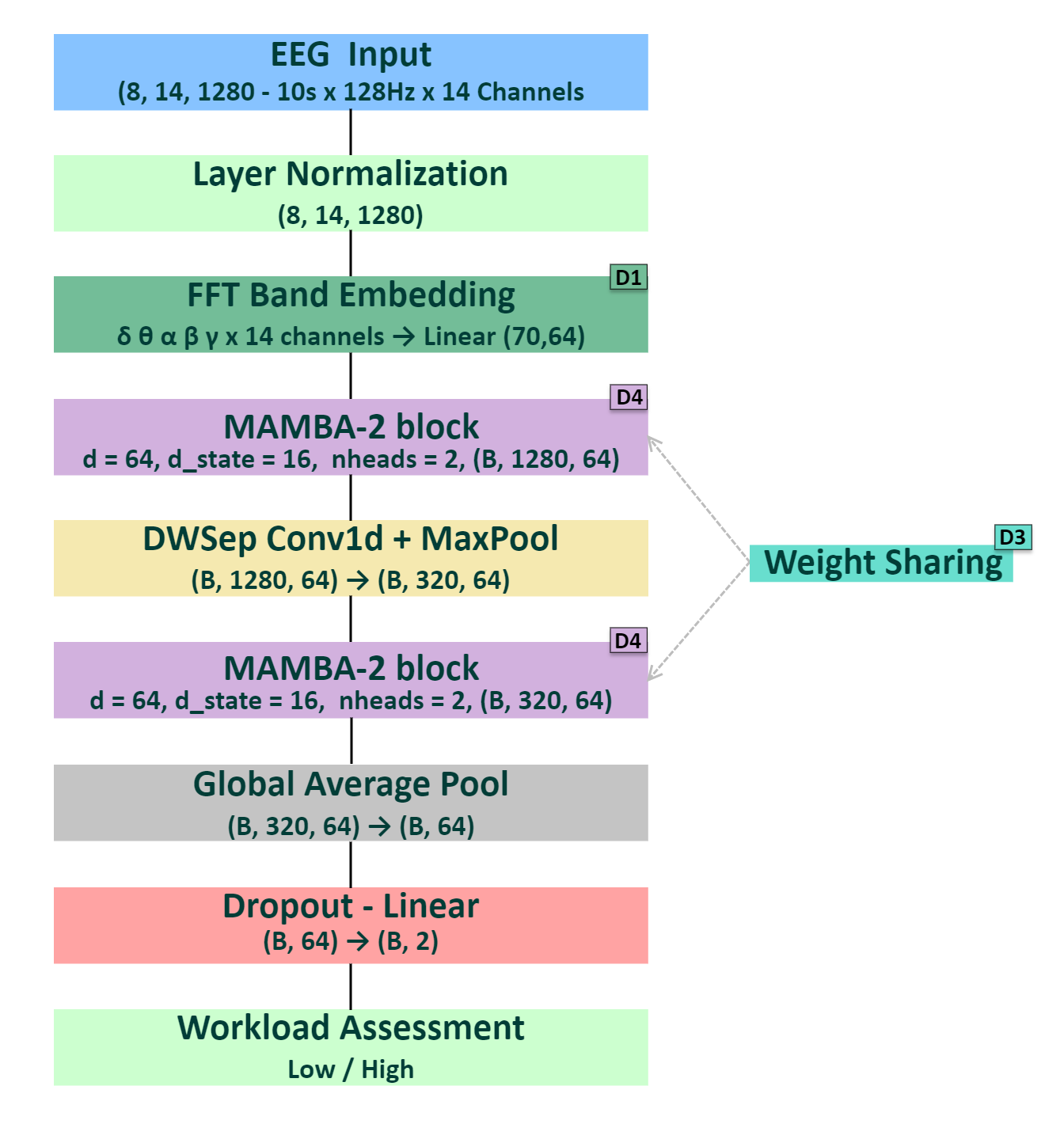}
    \caption{Overview of the proposed pipeline.}
    \label{fig:FOCUS}
\end{figure}

\subsubsection{Input Normalisation}
Each channel is independently normalized via Layer Normalization applied over the temporal axis:
\begin{equation}
  \hat{x}_c = \mathrm{LN}_T(x_c), \quad c = 1,\ldots,C,
  \label{eq:layernorm}
\end{equation}
so that the amplitude distribution of the data points is stable before any spectral decomposition is applied.

\subsubsection{D1: FFT Band Embedding}
The model does not learn filter banks from data. Instead, it decomposes each channel into five physiologically motivated frequency bands utilizing the discrete Fourier transform. Initially, each channel signal is multiplied by a Hann window in order to suppress spectral leakage:
\begin{equation}
  \tilde{x}_c(t) = x_c(t)\cdot w(t),\quad
  w(t)=\tfrac{1}{2}\!\left(1-\cos\!\frac{2\pi t}{T}\right),
  \label{eq:hann}
\end{equation}
and then transformed via the real-valued FFT:
\begin{equation}
  X_c(k) = \sum_{t=0}^{T-1} \tilde{x}_c(t)\,
            e^{-j2\pi kt/T},\quad k=0,\ldots,\tfrac{T}{2}.
  \label{eq:rfft}
\end{equation}
Next, five band-pass masks $\{M_b(k)\}_{b=1}^{5}$ are pre-computed with flat unit passbands and two-bin Hann-windowed roll-off transitions at each edge, and they cover the canonical EEG bands, namely $\delta$~(0.5--4\, Hz), $\theta$~(4--8\, Hz), $\alpha$~(8--13\, Hz), $\beta$~(13--30\, Hz), and $\gamma$~(30--43\, Hz). The band-filtered time-domain signal is recovered by the inverse FFT:
\begin{equation}
  x_c^{(b)}(t) = \mathcal{F}^{-1}\!\left\{
      X_c(k)\cdot M_b(k)\right\}(t).
  \label{eq:bandfilter}
\end{equation}
Afterward, the five-band signals of all channels are stacked and
reshaped into a sequence
$\mathbf{E} \in \mathbb{R}^{T \times 5C}$
(obtained by permuting the $(B,5,C,T)$ tensor to
$(B,T,5,C)$ and flattening the last two axes). A single
learned linear projection maps this sequence to the model dimension $d$:
\begin{equation}
  \mathbf{H}^{(0)} =
    \mathrm{LN}\!\left(\mathbf{E}\,\mathbf{W}_\mathrm{emb}^\top
    + \mathbf{b}_\mathrm{emb}\right) \in \mathbb{R}^{T \times d},
  \label{eq:fftproj}
\end{equation}
where $\mathbf{W}_\mathrm{emb} \in \mathbb{R}^{d \times 5C}$ is the only learnable component in the entire embedding stage, while the band masks have zero trainable parameters.
The inverse FFT in \eqref{eq:bandfilter} recovers each band signal at the \textit{full} temporal resolution of
$T=1280$ time points. The decomposition uses fixed frequency-domain bandpass masks, and no short-time Fourier transform is applied.
Consequently, the sequence $\mathbf{H}^{(0)}$ fed to the encoder has length $T$, not a coarser time-frequency grid.
At each time point $t$, the token $\mathbf{H}^{(0)}_t \in \mathbb{R}^d$ is a learned projection of the $5C=70$ band-filtered time-domain values $\{x_c^{(b)}(t)\}_{b,c}$. The Mamba-2 encoder therefore processes a sequence of $1280$ multi-band time-domain representations and learns how the band-wise activity of the full electrode array evolves over the $10$ seconds.
In total, the model has 0.04\, M parameters, and a forward pass costs 0.11\, GFLOPs. On an NVIDIA GeForce RTX 3090 GPU, we measured a model-forward latency of 0.08\, ms per sample, i.e., per complete 10\,s input of 14 EEG channels ($T=1280$ time points). The 10\,s needed to acquire the signal and the offline ICA artifact removal are not part of this value. It reflects the model computation alone and not the latency of the end-to-end decision.

\subsubsection{Mamba-2 Encoder Block (D4)}

The embedded sequence $\mathbf{H}^{(0)}$ is fed into a
Mamba-2 structured state-space model~\cite{mamba2}.
A latent state
$\mathbf{h}_t \in \mathbb{R}^{d_\mathrm{expand} \times N}$
is kept by the model for every time step $t$, and its evolution follows the discretized linear recurrence:
\begin{align}
  \mathbf{h}_t &= \bar{\mathbf{A}}_t\,\mathbf{h}_{t-1}
                 + \bar{\mathbf{B}}_t\,\mathbf{x}_t,
  \label{eq:ssm_state} \\
  \mathbf{y}_t &= \mathbf{C}_t\,\mathbf{h}_t,
  \label{eq:ssm_out}
\end{align}
where
$\bar{\mathbf{A}}_t = e^{\Delta_t \mathbf{A}}$ and
$\bar{\mathbf{B}}_t = \Delta_t\,\mathbf{B}_t$
are input-dependent, discretized matrices derived from the
learned step-size gate $\Delta_t$ and the selective context
matrices $\mathbf{B}_t,\mathbf{C}_t$. The state dimension is
$N = d_\mathrm{state} = 16$. With an internal expansion factor
of $\mathrm{expand}=2$, the block has
$d_\mathrm{expand} = 2d = 128$ inner channels.
A pre-norm residual connection wraps the SSM:
\begin{equation}
  \mathbf{H}^{(1)} =
    \mathbf{H}^{(0)}
    + \mathrm{SSM}_\mathrm{enc}\!\left(
        \mathrm{LN}\!\left(\mathbf{H}^{(0)}\right)\right)
    \in \mathbb{R}^{T \times d}.
  \label{eq:mamba_enc}
\end{equation}

\subsubsection{D2: Depthwise-Separable Temporal Convolution}

After the encoder, the sequence is convolved along the time axis
with a depthwise-separable convolution~\cite{kaiser2017depthwise}, in order to
capture local temporal patterns efficiently. With
$\mathbf{H}^{(1)} \in \mathbb{R}^{d \times T}$ written in
channels-first layout, the depthwise step applies one
independent kernel $\mathbf{k}_i \in \mathbb{R}^{1 \times k}$
per feature channel $i$ (kernel size $k=3$):
\begin{equation}
  h_i^{\mathrm{dw}}(t) =
    \sum_{\tau=0}^{k-1} k_i(\tau)\,H^{(1)}_i(t-\tau),
  \label{eq:depthwise}
\end{equation}
followed by Batch Normalization. The pointwise step then recombines
the channels via a $1{\times}1$ convolution and applies the activation:
\begin{equation}
  \mathbf{H}^{(2)} =
    \mathrm{GELU}\!\left(
      \mathrm{BN}\!\left(
        \mathbf{W}_\mathrm{pw}\,
        \mathrm{BN}\!\left(\mathbf{H}^{\mathrm{dw}}\right)
      \right)
    \right) \in \mathbb{R}^{d \times T}.
  \label{eq:pointwise}
\end{equation}
A standard convolution
$\mathbf{W}_\mathrm{std} \in \mathbb{R}^{d \times d \times k}$
requires $d^2 k$ parameters, while the depthwise-separable
factorisation uses $d k$ (depthwise) plus $d^2$ (pointwise)
$= d(k\!+\!d)$ parameters. This is a
$d^2k\,/\,d(k\!+\!d)= dk/(k+d)$ reduction. For
$d=64$ and $k=3$, it yields $12{,}288$ versus $4{,}288$
parameters, i.e., a $2.9\times$ saving.
Afterward, the temporal resolution is reduced by a factor of four via
max-pooling:
\begin{equation}
  \mathbf{H}^{(3)} =
    \mathrm{MaxPool}_{k=4}\!\left(\mathbf{H}^{(2)}\right)
    \in \mathbb{R}^{d \times \lfloor T/4 \rfloor}.
  \label{eq:maxpool}
\end{equation}

\subsubsection{D3: Bottleneck Block with Weight Sharing}

The down-sampled sequence
$(\mathbf{H}^{(3)})^\top \in \mathbb{R}^{T/4 \times d}$
is further refined by a \textit{bottleneck} Mamba-2 pass.
According to the weight-sharing decision (D3), this pass reuses the \textit{same}
parameter set as the encoder SSM:
\begin{equation}
  \mathbf{H}^{(4)} =
    \mathbf{H}^{(3)\top}
    + \mathrm{SSM}_\mathrm{enc}\!\left(
        \mathrm{LN}\!\left(\mathbf{H}^{(3)\top}\right)\right)
    \in \mathbb{R}^{T/4 \times d},
  \label{eq:mamba_bottleneck}
\end{equation}
where $\mathrm{SSM}_\mathrm{enc}$ is the identical module from
\eqref{eq:mamba_enc}. Weight sharing halves the SSM parameter
budget. For $d=64$, it saves $\approx\!28{,}000$
parameters.

\subsubsection{Classification Head}

A global temporal average pools the bottleneck output into a single descriptor, which is regularised via dropout ($p=0.3$) and mapped to class logits by a linear layer:
\begin{equation}
  \hat{y} = \mathbf{W}_\mathrm{cls}\,
    \mathrm{Dropout}\!\left(
      \frac{4}{T}\sum_{t=1}^{T/4}\mathbf{H}^{(4)}_t
    \right) \in \mathbb{R}^{2}.
  \label{eq:classifier}
\end{equation}

\subsection{Training Configuration}

All models are optimized with AdamW ($\eta\!=\!10^{-4}$, weight decay $\lambda\!=\!0.05$) under a
cosine annealing learning-rate schedule with $20$ warm-up and $10$ cool-down epochs over $190$ total training epochs. The batch size is $32$, and label smoothing is set to $0.10$. During training, data augmentation is applied per channel; polarity inversion, additive Gaussian noise, and temporal masking (window size $15-30$\% of the sequence) are each applied with a probability drawn uniformly from $[0.01, 0.10]$. 
For each LOSO fold, the model is initialized and trained from scratch on 10 participants, while the remaining participant is held out for evaluation. The reported result of each fold is the maximum accuracy achieved on the held-out participant within the fixed 190-epoch training horizon, while the final performance is computed as the mean across all 11 folds.

\section{Experimental Evaluation \& Results}
\label{sec:results}

In this section, we present the experiments on cognitive workload assessment from EEG signals, conducted under the LOSO evaluation protocol with $11$ participants and three cognitive tasks (IQ, MATH, and GAME). The primary objective is to identify the FOCUS configuration that provides the best overall trade-off between classification performance and computational cost. In order to assess their individual contribution to classification performance and computational cost, we varied four design decisions: weight sharing 
between SSM blocks (D3), convolution type (D2), model width $d$ (D4), and SSM state dimension $d_\text{state}$ (D4). 
The binary labels correspond to the predefined easy and hard task-difficulty conditions and do not represent direct subjective workload measurements.
Table~\ref{tab:ablation} reports the results separately for IQ, MATH, and GAME, as well as for ALL, which denotes a separate LOSO experiment on the full dataset with samples from all three tasks. Regarding ALL, a single model is trained in each LOSO fold on the full data of the training participants, and the task identity is not provided as an input to the model. ALL is therefore not an arithmetic average of the three task-specific results. Results are reported as mean $\pm$ std over 11 LOSO folds.
Accuracy is computed as the overall proportion of correctly classified samples, while precision and F1-score are macro-averaged across the EASY and HARD classes, i.e., both classes have equal weight.
In the ALL setting, each architectural variant was compared with A1 through a two-sided Wilcoxon signed-rank test on the 11 paired participant-level accuracies ($\alpha=0.05$), while the rank-biserial correlation is reported as the effect-size measure. No correction for multiple comparisons was applied.

\textbf{Weight sharing (D3).} The baseline A1 (weight-shared encoder and bottleneck blocks) is compared with A2, which trains two independent Mamba-2 blocks. Without weight sharing, the number of parameters increases from 0.04\, M to 0.07\, M, while the GFLOPs stay at 0.11. The effect on per-task accuracy is uneven. A2 performs better on IQ (71.97\% vs 67.42\%), whereas A1 performs better on MATH (75.76\% vs 69.70\%); on GAME, the accuracy is similar (76.14\% vs 76.89\%). In the ALL setting, A1 achieves 74.75\% accuracy compared with 70.20\% for A2, a difference of 4.55 percentage points, which is not significant under the paired Wilcoxon signed-rank test ($W=12.5$, $p=0.083$), with a rank-biserial correlation of $r=0.62$.
The numbers of parameters in Table~\ref{tab:ablation} are reported in millions and rounded to two decimal places.

\textbf{Depthwise-separable convolution (D2).} Replacing the depthwise-separable convolution with a standard Conv1d 
(A3) raises the number of parameters from 0.04\, M to 0.05\, M and adds 0.02\, GFLOPs. The accuracy differences across the three tasks are small; IQ increases from 67.42\% to 69.70\%, MATH decreases from 75.76\% to 75.00\%, and GAME is unchanged.
For ALL, the gap between A1 and A3 is 0.13 percentage points and is not statistically significant ($W=17.0$, $p=0.888$, rank-biserial $r=0.06$). We keep the depthwise-separable factorization for its 2.9$\times$ reduction in convolutional parameters (from 12,288 to 4,288), and not because of an equivalence in accuracy, which has not been established.

\textbf{Model width (D4, $d$).} We compare three widths, namely $d=32$ (A4), $d=64$ (A1), and $d=128$ (A5). The narrowest model loses 5.68 percentage points of ALL accuracy (69.07\% instead of 74.75\%), although this difference is not significant ($W=10.0$, $p=0.075$, rank-biserial $r=0.64$). At $d=128$, the ALL accuracy is 75.00\%. This is only 0.25 percentage points above A1 and again not significant ($W=32.0$, $p=0.966$, rank-biserial $r=-0.03$), whereas the parameters grow from 0.04\, M to 0.14\, M and the GFLOPs from 0.11 to 0.40. Consequently, $d=64$ is retained on the basis of the observed trade-off between cost and performance.

\textbf{SSM state dimension (D4, $d_\text{state}$).} Halving $d_\text{state}$ from 16 to 8 (A6) yields an ALL accuracy of 70.45\%, compared with 74.75\% for A1, and the difference of 4.30 percentage points is not significant ($W=14.0$, $p=0.102$, rank-biserial $r=0.58$). Lower accuracy is also observed for MATH (70.08\% vs 75.76\%) and GAME (69.32\% vs 76.14\%). The results suggest a sensitivity to the SSM state dimension in the present configuration; however, they show neither a statistically significant performance difference nor a specific temporal mechanism.

\textbf{Final configuration.} We adopt A1 as the final FOCUS architecture, i.e., $d=64$, $d_\text{state}=16$, depthwise-separable convolution, and weight-shared encoder and bottleneck blocks. A1 achieves 74.75\% ALL accuracy with 0.04\, M parameters, 0.11\, GFLOPs, and a model-forward latency of 0.08\, ms per sample. A5 ($d=128$) attains a slightly higher ALL accuracy of 75.00\%. However, the difference from A1 is only 0.25 percentage points and is not statistically significant ($W=32.0$, $p=0.966$, rank-biserial $r=-0.03$), while A5 requires 0.14\, M parameters and 0.40\, GFLOPs. For this reason, A1 is selected based on the cost--performance tradeoff.

\begin{table*}[t]
\caption{Ablation study of the architectural configurations. A1 is selected based on the overall performance--cost trade-off.}
\label{tab:ablation}
\centering
\small
\setlength{\tabcolsep}{4pt}
\renewcommand{\arraystretch}{1.12}

\begin{tabular}{@{}llccccccccc@{}}
\toprule
 & & \multicolumn{4}{c}{Configuration}
   & \multicolumn{2}{c}{Cost}
   & \multicolumn{3}{c}{Performance (\%)} \\
\cmidrule(lr){3-6}
\cmidrule(lr){7-8}
\cmidrule(l){9-11}
Config & Task
& $d$ & $d_{\text{state}}$ & DWSep & Shared
& Params (M) & GFLOPs
& Accuracy & Precision & F1 \\
\midrule

\multirow{4}{*}{\shortstack[l]{A1\\(selected)}}
  & IQ   & 64 & 16 & \checkmark & \checkmark
  & \multirow{4}{*}{0.04}
  & \multirow{4}{*}{0.11}
  & $67.42{\scriptstyle\,\pm6.84}$
  & $69.33{\scriptstyle\,\pm7.97}$
  & $66.77{\scriptstyle\,\pm7.00}$ \\

  & MATH & 64 & 16 & \checkmark & \checkmark
  & &
  & $\mathbf{75.76}{\scriptstyle\,\pm7.91}$
  & $77.00{\scriptstyle\,\pm7.89}$
  & $\mathbf{75.43}{\scriptstyle\,\pm8.06}$ \\

  & GAME & 64 & 16 & \checkmark & \checkmark
  & &
  & $76.14{\scriptstyle\,\pm7.96}$
  & $\mathbf{78.83}{\scriptstyle\,\pm7.66}$
  & $75.43{\scriptstyle\,\pm8.39}$ \\

\cmidrule(lr){2-6}\cmidrule(l){9-11}
  & \textit{ALL} & 64 & 16 & \checkmark & \checkmark & &
  & $74.75{\scriptstyle\,\pm10.54}$
  & $76.29{\scriptstyle\,\pm10.00}$
  & $74.13{\scriptstyle\,\pm11.11}$ \\

\midrule

\multirow{4}{*}{\shortstack[l]{A2\\(no sharing)}}
  & IQ   & 64 & 16 & \checkmark & $\times$
  & \multirow{4}{*}{0.07}
  & \multirow{4}{*}{0.11}
  & $\mathbf{71.97}{\scriptstyle\,\pm6.90}$
  & $\mathbf{73.63}{\scriptstyle\,\pm6.85}$
  & $\mathbf{71.40}{\scriptstyle\,\pm7.16}$ \\

  & MATH & 64 & 16 & \checkmark & $\times$
  & &
  & $69.70{\scriptstyle\,\pm6.67}$
  & $72.54{\scriptstyle\,\pm6.07}$
  & $68.44{\scriptstyle\,\pm7.86}$ \\

  & GAME & 64 & 16 & \checkmark & $\times$
  & &
  & $\mathbf{76.89}{\scriptstyle\,\pm7.82}$
  & $77.80{\scriptstyle\,\pm7.81}$
  & $\mathbf{76.68}{\scriptstyle\,\pm7.91}$ \\

\cmidrule(lr){2-6}\cmidrule(l){9-11}
  & \textit{ALL} & 64 & 16 & \checkmark & $\times$ & &
  & $70.20{\scriptstyle\,\pm10.45}$
  & $69.10{\scriptstyle\,\pm16.16}$
  & $68.27{\scriptstyle\,\pm13.92}$ \\

\midrule

\multirow{4}{*}{\shortstack[l]{A3\\(std conv)}}
  & IQ   & 64 & 16 & $\times$ & \checkmark
  & \multirow{4}{*}{0.05}
  & \multirow{4}{*}{0.13}
  & $69.70{\scriptstyle\,\pm10.06}$
  & $73.46{\scriptstyle\,\pm9.14}$
  & $67.97{\scriptstyle\,\pm11.75}$ \\

  & MATH & 64 & 16 & $\times$ & \checkmark
  & &
  & $75.00{\scriptstyle\,\pm7.54}$
  & $\mathbf{77.66}{\scriptstyle\,\pm7.07}$
  & $74.22{\scriptstyle\,\pm8.16}$ \\

  & GAME & 64 & 16 & $\times$ & \checkmark
  & &
  & $76.14{\scriptstyle\,\pm11.53}$
  & $77.01{\scriptstyle\,\pm11.73}$
  & $75.92{\scriptstyle\,\pm11.63}$ \\

\cmidrule(lr){2-6}\cmidrule(l){9-11}
  & \textit{ALL} & 64 & 16 & $\times$ & \checkmark & &
  & $74.62{\scriptstyle\,\pm8.01}$
  & $\mathbf{76.33}{\scriptstyle\,\pm7.23}$
  & $73.93{\scriptstyle\,\pm8.93}$ \\

\midrule

\multirow{4}{*}{\shortstack[l]{A4\\($d=32$)}}
  & IQ   & 32 & 16 & \checkmark & \checkmark
  & \multirow{4}{*}{0.01}
  & \multirow{4}{*}{0.03}
  & $62.88{\scriptstyle\,\pm8.79}$
  & $65.52{\scriptstyle\,\pm14.62}$
  & $58.66{\scriptstyle\,\pm13.76}$ \\

  & MATH & 32 & 16 & \checkmark & \checkmark
  & &
  & $68.56{\scriptstyle\,\pm5.99}$
  & $70.82{\scriptstyle\,\pm6.72}$
  & $67.65{\scriptstyle\,\pm6.43}$ \\

  & GAME & 32 & 16 & \checkmark & \checkmark
  & &
  & $64.39{\scriptstyle\,\pm7.40}$
  & $66.43{\scriptstyle\,\pm14.04}$
  & $60.62{\scriptstyle\,\pm12.39}$ \\

\cmidrule(lr){2-6}\cmidrule(l){9-11}
  & \textit{ALL} & 32 & 16 & \checkmark & \checkmark & &
  & $69.07{\scriptstyle\,\pm4.59}$
  & $69.52{\scriptstyle\,\pm4.38}$
  & $68.83{\scriptstyle\,\pm4.78}$ \\

\midrule

\multirow{4}{*}{\shortstack[l]{A5\\($d=128$)}}
  & IQ   & 128 & 16 & \checkmark & \checkmark
  & \multirow{4}{*}{0.14}
  & \multirow{4}{*}{0.40}
  & $68.94{\scriptstyle\,\pm8.21}$
  & $70.39{\scriptstyle\,\pm7.18}$
  & $67.79{\scriptstyle\,\pm9.75}$ \\

  & MATH & 128 & 16 & \checkmark & \checkmark
  & &
  & $73.86{\scriptstyle\,\pm9.75}$
  & $76.85{\scriptstyle\,\pm8.91}$
  & $72.79{\scriptstyle\,\pm10.84}$ \\

  & GAME & 128 & 16 & \checkmark & \checkmark
  & &
  & $75.38{\scriptstyle\,\pm6.75}$
  & $76.33{\scriptstyle\,\pm7.34}$
  & $75.20{\scriptstyle\,\pm6.73}$ \\

\cmidrule(lr){2-6}\cmidrule(l){9-11}
  & \textit{ALL} & 128 & 16 & \checkmark & \checkmark & &
  & $\mathbf{75.00}{\scriptstyle\,\pm9.08}$
  & $75.76{\scriptstyle\,\pm8.90}$
  & $\mathbf{74.69}{\scriptstyle\,\pm9.39}$ \\

\midrule

\multirow{4}{*}{\shortstack[l]{A6\\($d_{\text{state}}=8$)}}
  & IQ   & 64 & 8 & \checkmark & \checkmark
  & \multirow{4}{*}{0.04}
  & \multirow{4}{*}{0.11}
  & $63.26{\scriptstyle\,\pm10.78}$
  & $62.25{\scriptstyle\,\pm15.11}$
  & $61.52{\scriptstyle\,\pm13.53}$ \\

  & MATH & 64 & 8 & \checkmark & \checkmark
  & &
  & $70.08{\scriptstyle\,\pm7.91}$
  & $71.04{\scriptstyle\,\pm7.62}$
  & $69.61{\scriptstyle\,\pm8.33}$ \\

  & GAME & 64 & 8 & \checkmark & \checkmark
  & &
  & $69.32{\scriptstyle\,\pm7.80}$
  & $71.04{\scriptstyle\,\pm7.71}$
  & $68.60{\scriptstyle\,\pm8.12}$ \\

\cmidrule(lr){2-6}\cmidrule(l){9-11}
  & \textit{ALL} & 64 & 8 & \checkmark & \checkmark & &
  & $70.45{\scriptstyle\,\pm8.92}$
  & $71.85{\scriptstyle\,\pm8.19}$
  & $69.32{\scriptstyle\,\pm10.57}$ \\

\bottomrule
\end{tabular}
\end{table*}

\section{Comparison with Existing Work}
\label{sec:comparison}

Table~\ref{tab:comparison} compares FOCUS with three published references, EEGNet~\cite{lawhern2018}, 1BT~\cite{gkikas_cruz_eeite_cwl_2026}, and the multimodal framework proposed in~\cite{gkikas_cruz_acii_cwl_2026}, as well as with a plain Mamba-2 baseline evaluated in the present study. EEGNet and 1BT provide CNN- and Transformer-based EEG reference architectures, respectively, while the framework in~\cite{gkikas_cruz_acii_cwl_2026} employs five modalities, namely ECG, EDA, RESP, SpO$_2$, and EEG. The plain Mamba-2 baseline serves as a direct state-space-model comparison, in order to assess the overall benefit of the proposed FOCUS design. The published EEGNet and 1BT results were obtained on the same EEG data under the same subject-independent LOSO protocol, while the multimodal framework employs the same participants and task protocol with the additional peripheral modalities. Plain Mamba-2 and FOCUS were evaluated in the present study. As established in~\cite{gkikas_cruz_eeite_cwl_2026}, the mean score is the unweighted average across all three tasks and all three evaluation metrics.

FOCUS attains the highest accuracy of all evaluated models on MATH and GAME. Its MATH accuracy is 75.76\%, while plain Mamba-2 reaches 67.42\%, 1BT 67.05\%, the multimodal framework 64.39\%, and EEGNet 63.64\%. On GAME, the 76.14\% of FOCUS is followed by 70.83\% for the multimodal framework, 69.70\% for 1BT, 68.56\% for Mamba-2, and 64.39\% for EEGNet. IQ shows a different pattern. Here, the multimodal framework achieves the highest accuracy of 72.35\%, followed by 1BT at 69.32\%, Mamba-2 at 68.56\%, FOCUS at 67.42\%, and EEGNet at 65.15\%. The advantage of FOCUS is therefore observed primarily on MATH and GAME, while on IQ the multimodal framework achieves the highest performance. In the pooled ALL setting, FOCUS achieves 74.75\% accuracy, compared with 62.88\% for the multimodal framework, while the measured GPU model-forward latency of FOCUS is 0.08\, ms per sample.

The five models exhibit different cost--performance profiles. As the following comparison indicates, FOCUS provides a good trade-off between predictive performance and computational cost, and it does not minimize the computational cost alone. EEGNet is the most computationally compact, with 0.01\, M parameters and 0.02\, GFLOPs, but it has the lowest mean performance of 64.47\%. 1BT achieves a mean score of 69.37\% with 0.45\,M parameters and 0.02\,GFLOPs, while the multimodal framework reaches 69.77\% with 5.61\,M parameters and 0.36\,GFLOPs. Plain Mamba-2 achieves 68.78\% with 0.06\, M parameters and 0.14\, GFLOPs, whereas FOCUS attains the highest mean score of 73.57\% with 0.04\, M parameters and 0.11\, GFLOPs. Compared with plain Mamba-2, the mean score of FOCUS is 4.79 percentage points higher, and FOCUS utilizes approximately 33\% fewer parameters and 21\% fewer GFLOPs. Regarding 1BT, FOCUS has a 4.20 percentage-point higher mean score and approximately 11$\times$ fewer parameters, although 1BT has lower GFLOPs. In addition, FOCUS achieves a 3.80 percentage-point higher mean score than the multimodal framework and requires considerably fewer parameters and GFLOPs.

\section{Discussion \& Conclusion}

The final configuration was selected through an ablation across four design dimensions. Regarding weight sharing, the effect depends on the task. A2 achieves higher IQ accuracy than A1, A1 achieves higher MATH accuracy than A2, and GAME performance is similar. For ALL, A1 reaches 74.75\% and A2 70.20\%. A1 is thus 4.55 percentage points more accurate and also has fewer parameters, although the paired Wilcoxon signed-rank test does not find the difference significant ($W=12.5$, $p=0.083$, rank-biserial $r=0.62$). Weight sharing is therefore not shown to be a statistically significant performance improvement, and no implicit-regularisation mechanism is shown either, so we retain the shared design based on the observed ALL performance and its parameter efficiency.

Compared with the standard Conv1d, the depthwise-separable convolution has $2.9\times$ fewer convolutional parameters. The ALL accuracies of A1 and A3 are only 0.13 percentage points apart, and the difference is not statistically significant ($W=17.0$, $p=0.888$, rank-biserial $r=0.06$). The depthwise-separable design is kept for its parameter efficiency, while we do not claim that the two designs are equivalent in accuracy, nor that a specific separability mechanism is at work.
With $d_\text{state}$ reduced from $16$ to $8$, the ALL accuracy is 4.30 percentage points lower than that of A1, although the difference is not significant ($W=14.0$, $p=0.102$, rank-biserial $r=0.58$). Increasing the model width from $d=64$ to $d=128$ yields only a 0.25 percentage-point higher ALL accuracy, which is not significant ($W=32.0$, $p=0.966$, rank-biserial $r=-0.03$), and the computational cost increases considerably. For these reasons, the compact A1 configuration is selected based on the observed cost--performance tradeoff, and not on statistically established performance differences.
A1 is selected as the final FOCUS configuration, since it provides the best overall performance--cost trade-off among the evaluated variants, with 74.75\% ALL accuracy, 0.04\, M parameters, and 0.11\, GFLOPs. A5 attains a slightly higher ALL accuracy of 75.00\%. However, this difference of 0.25 percentage points is not statistically significant ($p=0.966$), and A5 requires a considerably greater computational cost (0.14\, M parameters and 0.40\, GFLOPs). The selected A1 configuration has a measured GPU model-forward latency of $0.08$\, ms per sample.
 
The study has two main limitations; the dataset comprises $11$ participants recorded in a single controlled laboratory environment, and the generalisability of the findings to clinical or operational settings with different EEG headsets or longer recording sessions has not been tested. For this reason, the present results are pilot evidence that the proposed architecture is workable and computationally efficient, and not evidence of broad population-level generalization.
The binary targets represent controlled easy/hard task-difficulty conditions, which were used to induce differences in cognitive demand; therefore, the reported results are not a direct estimation of subjective workload.
In addition, the present study does not isolate the contribution of the fixed FFT decomposition relative to alternative EEG front-ends. The FFT stage is thus a parameter-free, physiologically motivated design choice, and an improvement over alternative representations has not been established empirically. 
For future studies, we suggest a direct comparison of the fixed FFT decomposition with time-domain EEG without fixed band decomposition, learned temporal filters, wavelet-based representations, and learnable filterbanks. FOCUS should also be evaluated on larger cohorts with more demographic diversity.
The present study intentionally focuses on EEG-only modeling in order to isolate the efficiency of the proposed architecture. Extending FOCUS to multimodal physiological sensing is a relevant direction for future work. Especially for social and companion robots in educational or healthcare settings \cite{montiel_cruz_gkikas_eeite_2026}, recognition accuracy, model size, and inference time are all of great importance for practical deployment \cite{gkikas_cruz_pain_eeite_2026}.

\begin{table}[h]
\caption{Comparison of FOCUS with published baselines and a plain Mamba-2 baseline evaluated in the present study, on the same dataset and under the same subject-independent LOSO protocol.}
\label{tab:comparison}
\setlength{\tabcolsep}{3pt}
\renewcommand{\arraystretch}{1.1}

\resizebox{\columnwidth}{!}{%
\begin{tabular}{@{}llcc
S[table-format=2.2(2)]
S[table-format=2.2(2)]
S[table-format=2.2(2)]@{}}
\toprule
Model & Task & \makecell{Params\\(M)} & GFLOPs
      & {\makecell{Acc\\(\%)}}
      & {\makecell{Prec\\(\%)}}
      & {\makecell{F1\\(\%)}} \\
\midrule

\multirow{4}{*}{EEGNet~\cite{lawhern2018}}
  & IQ   & \multirow{3}{*}{0.01} & \multirow{3}{*}{0.02}
  & 65.15 +- 6.5 & 65.90 +- 6.6 & 64.75 +- 6.7 \\
  & MATH & & &
  63.64 +- 13.7 & 63.95 +- 13.9 & 63.32 +- 13.8 \\
  & GAME & & &
  64.39 +- 9.5 & 64.93 +- 9.9 & 64.19 +- 9.4 \\
\cmidrule(l){2-7}
  & mean$^{*}$ & & & \multicolumn{3}{c}{64.47} \\

\midrule

\multirow{4}{*}{1BT~\cite{gkikas_cruz_eeite_cwl_2026}}
  & IQ   & \multirow{3}{*}{0.45} & \multirow{3}{*}{0.02}
  & 69.32 +- 7.8 & 71.22 +- 8.6 & 68.55 +- 8.2 \\
  & MATH & & &
  67.05 +- 7.6 & 72.73 +- 7.3 & 64.43 +- 10.2 \\
  & GAME & & &
  69.70 +- 8.9 & 73.65 +- 6.7 & 67.70 +- 11.6 \\
\cmidrule(l){2-7}
  & mean$^{*}$ & & & \multicolumn{3}{c}{69.37} \\

\midrule

\multirow{4}{*}{Unified~\cite{gkikas_cruz_acii_cwl_2026}}
  & IQ   & \multirow{3}{*}{5.61} & \multirow{3}{*}{0.36}
  & \multicolumn{1}{c}{\textbf{72.35}$^{\dagger}$}
  & \multicolumn{1}{c}{\textbf{76.27}$^{\dagger}$}
  & \multicolumn{1}{c}{\textbf{70.43}$^{\dagger}$} \\
  & MATH & & &
  \multicolumn{1}{c}{64.39$^{\dagger}$}
  & \multicolumn{1}{c}{69.00$^{\dagger}$}
  & \multicolumn{1}{c}{62.06$^{\dagger}$} \\
  & GAME & & &
  \multicolumn{1}{c}{70.83$^{\dagger}$}
  & \multicolumn{1}{c}{72.62$^{\dagger}$}
  & \multicolumn{1}{c}{69.99$^{\dagger}$} \\
\cmidrule(l){2-7}
  & mean$^{*}$ & & & \multicolumn{3}{c}{69.77} \\

\midrule

\multirow{4}{*}{Mamba-2}
  & IQ   & \multirow{3}{*}{0.06} & \multirow{3}{*}{0.14}
  & 68.56 +- 7.82 & 71.28 +- 7.35 & 67.30 +- 9.28 \\
  & MATH & & &
  67.42 +- 7.71 & 71.77 +- 6.96 & 65.35 +- 10.17 \\
  & GAME & & &
  68.56 +- 11.97 & 72.30 +- 10.95 & 66.50 +- 14.19 \\
\cmidrule(l){2-7}
  & mean$^{*}$ & & & \multicolumn{3}{c}{68.78} \\

\midrule

\multirow{4}{*}{FOCUS (ours)}
  & IQ   & \multirow{3}{*}{0.04} & \multirow{3}{*}{0.11}
  & 67.42 +- 6.8 & 69.33 +- 8.0 & 66.77 +- 7.0 \\
  & MATH & & &
    \bfseries 75.76 +- 7.9
    & \bfseries 77.00 +- 7.9
    & \bfseries 75.43 +- 8.1 \\
    
    & GAME & & &
    \bfseries 76.14 +- 8.0
    & \bfseries 78.83 +- 7.7
    & \bfseries 75.43 +- 8.4 \\
\cmidrule(l){2-7}
  & mean$^{*}$ & & & \multicolumn{3}{c}{\textbf{73.57}} \\

\bottomrule
\end{tabular}%
}

\begin{tablenotes}[para,flushleft]
\scriptsize
$^{*}$mean: unweighted average across IQ, MATH, and GAME and the three reported metrics.
$^{\dagger}$Standard deviation was not reported for the task-specific results.
\end{tablenotes}
\end{table}

\section*{Acknowledgments}
The authors used large language model (LLM)-based tools for language editing and improvement. All scientific content, results, and conclusions are solely the work of the authors.

\bibliographystyle{IEEEtran}
\bibliography{library}

\end{document}